# Jointly spatial-temporal representation learning for individual trajectories

Fei Huang [a,b], Jianrong Lv [a,b], Yang Yue [a,b,*]

[a] *Department of Urban Informatics, School of Architecture and Urban Planning, Shenzhen University, China*

[b] *Shenzhen Key Laboratory of Spatial Smart Sensing and Services, Shenzhen, China*

**ABSTRACT**

Individual trajectories, capturing significant human-environment interactions across space and time, serve as vital inputs for geospatial foundation models (GeoFMs). However, existing attempts at learning trajectory representations often encoded trajectory spatial-temporal relationships implicitly, which poses challenges in learning and representing spatiotemporal patterns accurately. Therefore, this paper proposes a joint spatial-temporal graph representation learning method (ST-GraphRL) to formalize structurally-explicit while learnable spatial-temporal dependencies into trajectory representations. The proposed ST-GraphRL consists of three compositions: (i) a weighted directed spatial-temporal graph to explicitly construct mobility interactions over space and time dimensions; (ii) a two-stage joint encoder (i.e., decoupling and fusion), to learn entangled spatial-temporal dependencies by independently decomposing and jointly aggregating features in space and time; (iii) a decoder guides ST-GraphRL to learn mobility regularities and randomness by simulating the spatial-temporal joint distributions of trajectories. Tested on three real-world human mobility datasets, the proposed ST-GraphRL outperformed all the baseline models in predicting movements' spatial-temporal distributions and preserving trajectory similarity with high spatial-temporal correlations. Furthermore, analyzing spatial-temporal features in latent space, it affirms that the ST-GraphRL can effectively capture underlying mobility patterns. The results may also provide insights into representation learnings of other geospatial data to achieve general-purpose data representations, promoting the progress of GeoFMs.

## 1. Introduction

The emergence of foundation models like GPT, Llama, and DALL-E, which are trained on large amounts of data and can be tailored to a range of downstream tasks, has triggered a paradigm shift in artificial intelligence (AI). Representation learning is regarded as a crucial technique for foundation models because it offers a unified vector representation of information to effectively capture and comprehend a subject matter (Bengio et al., 2013; Bommasani et al., 2022). Geospatial Foundation models, or GeoFMs, are also gaining increasing attention (Janowicz et al., 2020; Mai et al., 2023). Similarly, the effectiveness of GeoFMs is profoundly influenced by geospatial data representations (or features) that are implemented in learning geographical knowledge (Ge et al., 2022; Scheider & Richter, 2023). Most of the existing attempts, however, represent spatial and temporal information implicitly, resulting in the limited modelling capacity of GeoFMs (Mai et al., 2022; Zhang et al., 2023).

Individual trajectory data—typical human active data indicating interactions between humans and the environment over space and time—is regarded as a fundamental aspect of the study on human-environment relations (Hägerstrand, 1970; Liu & Biljecki, 2022; Murray et al., 2023). There being abundant coordinate points in trajectories, however, travel points in trajectories are usually sparse and provide limited activity information (Tian et al., 2022; Yao et al., 2023). Furthermore, individual trajectories sampled with irregular frequency contain a significant variety of structures. This complexity makes it challenging to model continuous time variables and discrete location variables in trajectories to learn latent spatial-temporal dependencies. Sequential structure models such as LSTM (Hochreiter & Schmidhuber, 1997) and Transformer (Vaswani et al., 2017) have been used to capture mobility location regularities hidden in the continuous time dimension while ignoring the structure information between non-consecutive locations (Jiang et al., 2022). Structured graphs leveraging deep learning models like GNN (Veličković et al., 2018) have recently been employed to discover high-order correlation relations in non-regular location data (Zhu et al., 2021), such as trajectories. Researchers in Geospatial artificial intelligence (GeoAI) have realized the importance of data representing, especially to the locations and space information in geospatial data (Liu et al., 2019; Mai et al., 2020; Niu & Silva, 2021); while the temporal aspect is also significant in representations (Mai et al., 2022). Each movement in trajectories presents a joint distribution over space and time (Schneider et al., 2013; Jiang et al., 2017), existing approaches still require enhancement to explicitly learn the spatial-temporal dependency to capture or infer spatial-temporal patterns accurately.

Therefore, this study proposes a joint spatial-temporal trajectory graph representation learning method (ST-GraphRL) to generate accurate and generalized individual trajectory representation via explicitly structured spatial-temporal information. The contributions of this work are two-fold:

- A novel directed spatial-temporal trajectory graph (ST-Graph) is constructed. The ST-Graph can explicitly present and extract the interdependent structural relations between locations and time concealed within sparse movements, overcoming structural design limitations in irregular trajectories with arbitrary lengths.
- A two-stage joint encoding framework is developed for individual trajectory representation (ST-GraphRL). This framework demonstrates proficiency in modelling spatial-temporal joint distributions and learning the intricate dependencies within trajectories, providing useful general-purpose representation for a wide range of downstream tasks on human mobility.

The remainder of this paper is organized as follows. Sections 2 reviews previous work on representation learning for trajectory, and Section 3 future gives definitions. In Section 4, we introduce the proposed ST-GraphRL model and its technical details. Section 5 reports on data processing and experimental results. We take discussions in Section 6. In the final section, we provide insights on future work in addition to a summary.

## 2. Related works

Representation learning, the process of acquiring low-dimensional general-purpose features or representations of data, is crucial for AI to discern and disentangle underlying explanatory factors from low-level sensory data (Bengio et al., 2013). This part reviews representation learning primarily through the lens of deep learning algorithms which is the dominant approach to representation learning at the current stage. Deep learning achieves representation learning by composing multiple non-linear transformations, automatically yielding representations applicable to various downstream tasks, spanning recognition, classification, and prediction (Grill et al., 2020). It spans across diverse data modalities including images (Caron et al., 2020), text (Devlin et al., 2019), and graphs (Veličković, et al., 2018). As typical spatial-temporal data, trajectory consists of consecutive (from a sequence) or spatially nearby observations. Consequently, the purpose of trajectory representation learning is to transform and compress the spatial-temporal distribution of raw trajectories into generic low-dimensional representation vectors, empowering the downstream tasks such as trajectory similarity computation (Glake et al., 2022), traffic prediction (Zhao et al., 2022), and urban structure analysis (Zhu et al., 2020). In trajectory representation learning, a trajectory can take two forms: as a sequence or a structured graph (Barbosa et al., 2018), either way, both temporal and spatial features are two indispensable factors that should be embedded into the trajectory representation (Damiani et al., 2020).

To handle the consecutive temporal dimension, trajectories are usually fed into a model as the ordered coordinates. Researches, such as t2vec (Li et al., 2018) considered trajectories as sequences of GPS sample points and learned the representation by reconstructing high-sampling trajectory sequences from low-sampling sequences. However, the reconstruction process did not match the temporal information under the sequential time order. As to model the feature relation between time $t$ and $t$+1, both LSTM-based and Transformer (Self-attention-based) models are feasible approaches that can capture the high-order correlations in mobilities. Although LSTM has been utilized to extract user mobility patterns from the check-in data (Gao et al., 2017; Kong & Wu, 2018), it is susceptible to disturbances caused by sparse and discontinuous data, which further complicates the identification of dependencies in movements (Glake et al., 2022). The Self-attention-based model enhances the ability to capture contextual features by dynamically allocating attention weights to different locations in the input trajectory over time; however, learning spatial transit relationships in moving processes remains challenging (Jiang et al., 2022; Fang et al., 2022). Furthermore, directly feeding a series of trajectories into either LSTMs or Self-attention-based models for temporal dimensional learning is ineffective, since it fails to address temporal characteristics adequately.

Since spatial linkages are identifiable through the graphical structure, graph deep learning has become a popular method for processing trajectory data (Martin et al., 2023; Tao et al., 2022). In this way, the trajectory is regarded as a structured graph in which the visited locations are represented as nodes and the mobilities are represented as edges. Such as user activity graphs and spatial knowledge graphs have been constructed to quantify individuals' mobility characteristics (Wang et al., 2019; Wang & Zhu, 2024). Generally, road networks possess the inherent advantage in describing spatial activity relationships, thus, road representations can be converted into trajectories through marked locations and then separately training sequence learning model with a self-supervised task to learn trajectory representations (Chen et al., 2021;

Yang et al., 2021). However, this kind of method separated space or time dimension, treating space and time distribution, i.e., $p(s)$ and $p(t)$, as two completely independent variables, which overlooks the spatial-temporal interactions in trajectories. Therefore, researchers have attempted to integrate structural information with sequential information and extract their dependent relationships to learn coupled spatial-temporal features (Yang et al., 2020; Gao et al., 2022; Liu et al., 2023; Yu & Wang, 2023). In which the GNN usually be used for the spatial structures embedding by capturing the long-range correlation between travels, then a sequential-structure model such as LSTM network or Self-attention-based model following to independently model temporal correlations in trajectories. However, these methods are to decompose the spatial-temporal distribution into conditionally dependent distributions, i.e., $p(t)$ and $p(s|t)$, in which spatial relationships are presented explicitly, while temporal information as an implicit condition in calculation. That may lead to restricted learning of spatial-temporal correlations. Additionally, the existing work based on spatial-temporal graph, such as integrated space and time dimensions of mobility data into node features (Liu et al., 2022), is hindered by complex and ambiguous spatial-temporal relationships in movements.

A review of prior research reveals a research gap regarding the joint representation of space and time, particularly given the difficulty of explicitly representing the temporal dimension. Thus, this study developed the ST-Graph to encode both spatial and temporal aspects while learning the joint spatial-temporal distribution of trajectories.

## 3. Method

This study proposes a spatial-temporal graph representation learning (ST-GraphRL) framework to learn the generalized low-dimensional vector that represents individuals' mobility pattern over space and time. Formally, we aim to find a mapping function $f: \mathcal{G} \rightarrow \mathcal{H}$, which takes the trajectory graph $\mathcal{G}$ as the input, and outputs the vectorized representation $\mathcal{H}$ of individual that is subject to the constraints of the mobility distribution. The proposed method, as shown in Figure 1, consists of two essential parts: (i) constructing an individual spatial-temporal trajectory graph $\mathcal{G}$ to organize the individual's mobility, in which the node refers to the spatial attribute, the edge linking with two nodes refers to the temporal attribute in a complete movement; (ii) developing a GNN-based representation learning framework to learn low-dimensional vectorial representation of mobility. It is composed of a two-stage jointly encoder and a decoder that learn mobility distributions and patterns through decoupling and fusing spatial-temporal dependency.

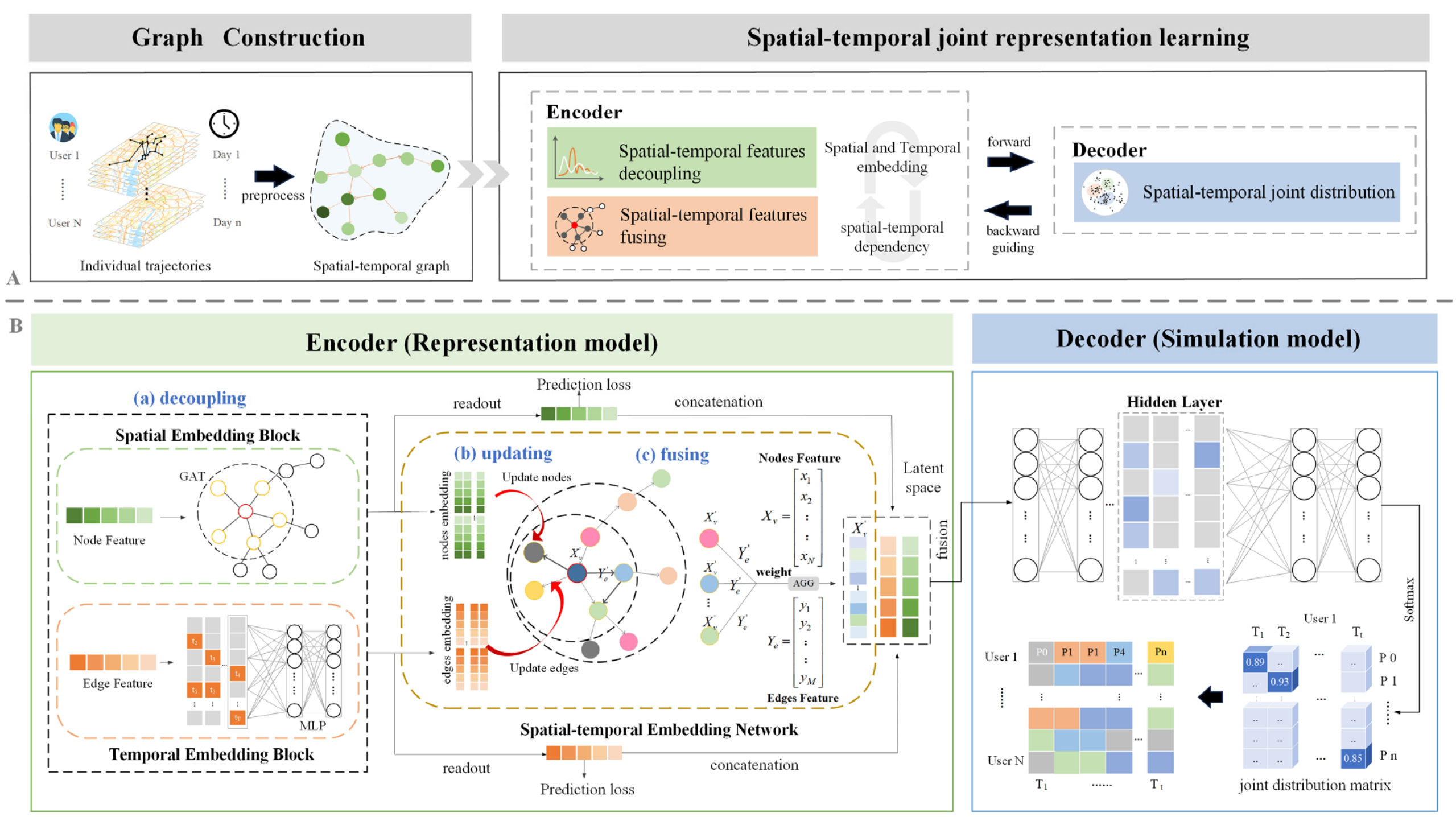

**Figure 1.** Workflow of ST-GraphRL; subfigure A presents the workflow of ST-GraphRL, subfigure B presents details of the encoder and decoder in ST-GraphRL.

## 3.1 Spatial-temporal trajectory graph

**Definition of individual spatial-temporal trajectory graph.** Given a trajectory sequence set $\mathcal{R}(u) = (S_1, S_2, \dots, S_\mathcal{T})$ of user $u \subset U$, in which $S_\mathcal{T} = (l_1, \dots, l_i)$ refers to the collected trajectory with $i$ visited locations in the $\mathcal{T}$-th day. Individuals' trajectories are constructed as a graph $\mathcal{R}(u) \rightarrow \mathcal{G} = (\mathcal{V}, \mathcal{E}, \mathcal{W})$, where $\mathcal{V} =< v_j^{1\times d_s} >_{j=1}^{|\mathcal{V}|}$ is the nodes set, and $\mathcal{E} =< e_j^{1\times d_t} >_{j=1}^{|\mathcal{V}|-1}$ is the edge set; $d_s$=130 and $d_t$=48 refer to the length of vectors in $\mathcal{V}$ and $\mathcal{E}$, respectively. The nodes on the graph are made up of geographical positions (longitude, latitude), and semantic attributes denoted as (location attributes, category attributes). Thus, the node is structed as $v_j = (lat_j, lon_j, cat_j)$ i.e., a pair of latitude and longitude, and the category or semantic features of the location. The edge records the dynamic temporal variation in a complete mobility, in which the "hourly" attribute was encoded into 48-dimensional binary vector linking two spatial locations. Thereby, a complete mobility is presented as $(v_j, e_j, v_{j+1})$ . Here, assuming $e_j = [0,0,1,0,0,0,1, \dots \dots]$, it indicates that the user left from location $j$ at 1:30am and then arrived at location $j$+1 at 3:30am. The edges' weight $\mathcal{W} = (f, dis, dura)$ contains the visit frequency, distance, and duration between two locations.

The conventional approaches construct graphs by combining the spatial-related information such as locations and POI categories with time information to form node features, which overlooks features in mobility transfers and breaks spatial-temporal interactions in movements. Movement is an entangled spatial-temporal transfer process, where the transition of locations and the accompanying temporal changes need to be explicitly structured. The structure of $(v_j, e_j, v_{j+1})$ with the edges' weight $\mathcal{W}$ explicitly presents the relationships between space and time in the mobility, which is helpful for the deep learning model to learn the spatial-temporal dependency and furtherly infer mobility patterns.

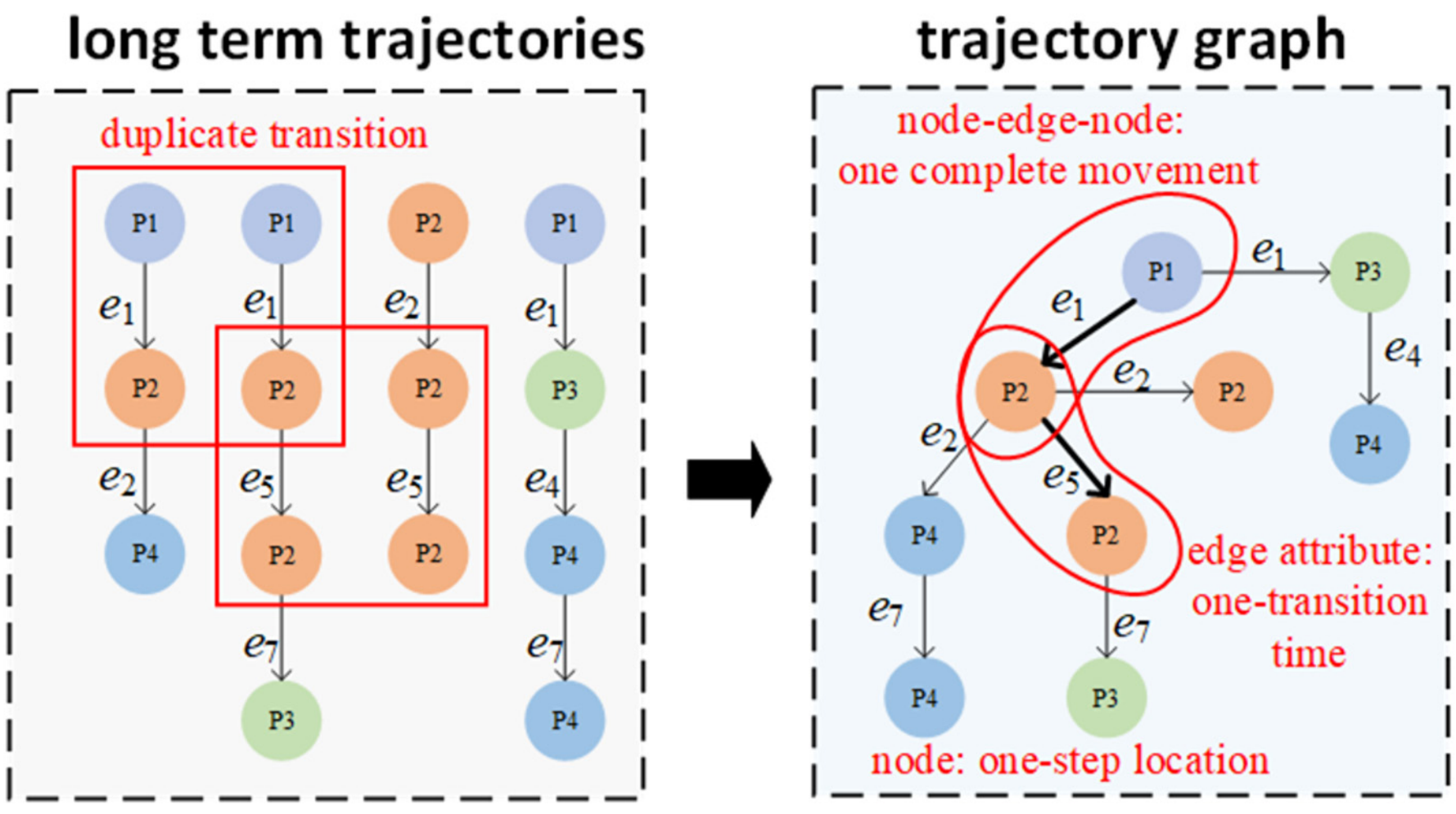

**Figure 2.** Details of the construction to the spatial-temporal trajectory graph of individual. one complete movement, presented in the form of node-edge-node, includes origin-destination and start-end time; for example, the user left from place P1 at 8:30 am and arrived at place P2 at 10:00 am, where $e_i$ indicates its transit time i.e., from 8:30am to 10:00am. duplicate transition refers to two or more identical complete movements, and its identified mobility frequency is visually reflected by the thickness of the edge.

As the example shown in the Figure 2, lengths of trajectories collected in the long term are not consistent and the information in a trajectory is sparse, up to four movement points. $\mathcal{G}$ builds the mobility regularity and randomness across multiple trajectories by utilizing these transit attributes (i.e., $e_i$ and duplicate transition) as the linking bridge. In this example, since the individual is more likely to move from P1 to P2 than to P3 in time-period $e_1$, as well as preferring moving from P2 to P2 in time-period $e_5$ than in $e_2$. (see the red rectangles in the trajectory graph). Therefore, $\mathcal{G}$ denotes P1-$e_1$-P2 and P2-$e_5$-P2 with a higher wight (marked by the red circles) than P1-$e_1$-P3 and P2-$e_2$-P2, which captures the temporal and spatial routines in travel transitions, meanwhile preserving randomness in movements. It is found that $\mathcal{G}$ describes a travel as an explicit transition in both time and location, where time acts as a significant mediator between the shifting location information, such that explicitly building the intricate spatial-temporal interaction from irregular trajectories with sparse information.

### 3.2 Representation module

To effectively represent the coupling correlations between space and time in trajectories, we designed a two-stage encoder. It models the entangled space and time features by decomposing the learning into two steps: decoupling and fusing. As shown in Figure 1 B-(a), the encoder first decomposes spatial-temporal information by the spatial and temporal embedding block, respectively. It aims to model the independent feature distributions, i.e., $p(s)$ and $p(t)$, providing initial awareness of space and time features. These decoupled features are furtherly embedded into graph $\mathcal{G}$ and update its nodes and edges, which are then fused to capture low-dimensional spatial-temporal dependencies, i.e., $p(s_{i+1}|s_i, t_i)$ and $p(t_{i+1}|s_i, t_i)$ through the message passing and aggregation operation of the spatial-temporal network, as shown in Figure 1 B-(b-c).

#### 3.2.1 Decoupling spatial-temporal information

Considering the mobility distribution, there, $P(\Phi_s|\mathcal{V})$, $P(\Phi_t|\mathcal{E})$, and $P\left(\Phi_{t,s}\middle|\mathcal{H}\right)$ denote the matrix of location categories distributions, travel timing distributions, and spatial-temporal distributions of trajectories in $u$, respectively; in which $\Phi_s \subset \mathbb{R}^{U\times d_s}$, $\Phi_t \subset \mathbb{R}^{U\times d_t}$, $\Phi_{t,s} \subset \mathbb{R}^{U\times d_s \times d_t}$, and $\mathcal{H}$ indicates the latent representation of $\mathcal{G}$.

In the spatial-temporal information decoupled modelling, $\mathcal{V}$ and $\mathcal{E}$ refer to space and time information decoupled from $\mathcal{G}$, respectively, in which $\mathcal{V}$ and $\mathcal{E}$ are fully independent. Spatial embedding block works based on a GAT model (Veličković et al., 2018), which helps to learn the topology of locations in mobility trajectory, hence $\hat{\mathcal{V}} = f_{GAT}(\mathcal{V})$. Temporal embedding block relies on a stack of multilayer perceptron (MLP) units, in which $\hat{\mathcal{E}} = f_{MLP}(\mathcal{E})$. Furtherly, distributions $P(\Phi_s|\mathcal{V})$ and $P(\Phi_t|\mathcal{E})$ are regarded as the guidance of training for $\hat{\mathcal{V}}$ and $\hat{\mathcal{E}}$, such that the preliminary embeddings containing completely latent regularities in space and time can be achieved. Here, $\Phi_s = f_{sq-s}(meanpool(\hat{\mathcal{V}}))$, and $\Phi_t = f_{sq-t}(sum(w_t * \hat{\mathcal{E}}))$, $w_t$ denotes a trainable weight for time information, $f_{sq-s}$ and $f_{sq-t}$ refer to the linear mapping for spatial embeddings and temporal embeddings.

#### 3.2.2 Learning spatial-temporal dependency

Learning spatial-temporal joint embedding for trajectories, which involves estimating the probability distribution $p(s,t)$ linked to time $t$ and location $s$, is intractable but crucial for capturing spatial-temporal patterns of movements. However, most previous researches have modeled it under the assumption of conditional independence. For example, the general approach extracts the temporal distribution $p(t)$, and then the conditional distribution $p(s|t)$ for spatial information. The information derived from the representation learned through $p(s|t)$ is implicit and constrained, revealing reduced expressiveness. Thus, we design jointly encoding to fuse the well-embedded space and time features in section 3.2.1, and learn spatial-temporal dependencies. In this part, $\mathcal{G}$ is updated to $\hat{\mathcal{G}} = (\hat{\mathcal{V}}, \hat{\mathcal{E}}, \mathcal{W})$, then a generalized GNN model (Li et al., 2020) with three layers is employed to capture the interdependence in spatial and temporal domains, which facilitates the next learning of spatial-temporal joint distributions. Detailly, the message updating of vertex $v$ is shown as:

$$m_{vu}^{l} = \rho^{l}\left(h_{v}^{l}, h_{u}^{l}, h_{e_{vu}}^{l}\right) = ReLu\left(h_{u}^{l} + h_{e_{vu}}^{l} + h_{w_{vu}}^{l}\right) + \epsilon \tag{1}$$

$$m_{e_{vu}}^{l} = MLP\Big(Concate(h_{v}^{l}, h_{u}^{l})\Big) \tag{2}$$

where $\zeta^{l}(\cdot)$ is the message aggregation function which outputs the aggregated message $m_{v}^{l}$; $h_{v}^{l}$ is the vertex features in the $l$-th layer, its neighbor's features $h_{u}^{l}$, the feature of edge attribute $h_{e_{vu}}^{l}$, and the normalized edge weight $h_{w_{vu}}^{l}$. There are two different aggregation operations for nodes and edges, written as:

$$\zeta_{v}^{l} = \sum\nolimits_{u \in \hat{\mathcal{V}}} \frac{exp(\beta m_{vu})}{\sum_{i \in \mathcal{N}(\hat{\mathcal{V}})} exp(\beta m_{vi})} \tag{3}$$

$$m_{v}^{l} = \zeta_{v}^{l}\left(\left\{m_{vu}^{l} \middle| u \in \hat{\mathcal{V}}\right\}\right) \tag{4}$$

$$h_{v}^{l+1} = \emptyset^{l}(h_{v}^{l}, m_{v}^{l}) = MLP\left(h_{v}^{l} + s \cdot \left\|h_{v}^{l}\right\|_{2} \cdot \frac{m_{v}^{l}}{\left\|m_{v}^{l}\right\|_{2}}\right) \tag{5}$$

$$h_{e_{vu}}^{l+1} = \emptyset^{l}\left(h_{e_{vu}}^{l}, m_{e_{vu}}^{l}\right) \tag{6}$$

As described above, the process of message passing, in which every updating for vertex and corresponding edge, seen as the joint estimation in $f(\hat{\mathcal{V}}^{l+1}|\hat{\mathcal{V}}^{l}, \hat{\mathcal{E}}^{l})$ and $f(\hat{\mathcal{E}}^{l+1}|\hat{\mathcal{V}}^{l}, \hat{\mathcal{E}}^{l})$, is the learning of interactive information in space and time. The representation $\mathcal{H}$ of the user is generated through the final aggregation:

$$a_{\mathcal{W}} = softmax\big(MLP(h_{\mathcal{W}})\big),\ a_{\mathcal{W}} \in \mathbb{R}^{(|\hat{\mathcal{V}}|-1)\times 1} \tag{7}$$

$$m_{\hat{\mathcal{V}}} = MLP\left(concat\left(h_{v}, h_{e_{vu}}, h_{u}\right)\right), (v, u) \in \hat{\mathcal{V}} \tag{8}$$

$$\mathcal{H} = sum(a_{\mathcal{W}} \cdot m_{\hat{\mathcal{V}}}), \mathcal{H} \in \mathbb{R}^{1\times d} \tag{9}$$

where d refers to the numbers of dimensions in $\mathcal{H}$, it is set $d = 24$.

### 3.3 Simulation module

To ensure the latent state $\mathcal{H}$ captures the pattern of spatial-temporal transition in long-term movements, we designed a simulation stage that can be regarded as a decoder to reconstruct the spatial-temporal distribution from $\mathcal{H}$. If the reconstruction is well accomplished, entangled spatial-temporal patterns can be learned for representation. The input of the simulation module is $z = (\mathcal{H}, \Phi_s, \Phi_t)$. As shown in Figure 1, several dense MLP units, here we stacked three units with residual structure, are utilized to explore the most possible location category and time for individual's movements.

$$\Phi_{t,s} = \mathcal{F}_{MPL}(z) \tag{10}$$

### 3.4 Loss function

We treat the training as the multi-label prediction problems to ensure that ST-GraphRL learns the accurate spatial-temporal distribution of mobility. The loss function includes: (i) Time distribution loss (in section 3.2.1) and (ii) Space distribution loss (in section 3.2.1), which aims to minimize the modeling error of capturing travel time/place preferences by simulating the distribution of time occurrences/visited places, guiding the model to learn temporal/spatial information embedded in user trajectories; and (iii) similarly, space-time joint distribution loss is designed to learn the spatial-temporal dependency (in section 3.2.2). Note that we employ the distribution balance loss (Wu et al., 2020) to overcome the bias (or imbalance) of travel time and visited places for individual trajectories and minimize the overall loss $L$ as follows. Here, $\hat{r}_i^k$ is a re-balancing weight, $z_i^k$ denotes the variation of $x^k$, $x_s = sigmoid(\Phi_s)$, $x_t = sigmoid(\Phi_t)$, and $x_{st} = softmax(\Phi_{t,s})$

$$\begin{aligned} L_{DB}(x^k, y^k) = \frac{1}{C}\sum_{i=0}^{C} \hat{r}_i^k \, [y_i^k \, log\left(1 + e^{-\left(z_i^k - v_i\right)}\right) \\ + \frac{1}{\lambda}\left(1 - y_i^k\right) log\left(1 + e^{-\left(z_i^k - v_i\right)}\right) \end{aligned} \tag{11}$$

$$L = 0.1 * L_{DB}(x_s, y_s) + \ 0.1 * L_{DB}(x_t, y_t) + L_{DB}(x_{st}, y_{st}) \tag{12}$$

## 4. Experiments and Result

### 4.1 Experimental Setup

#### 4.1.1 Datasets and preprocessing

Data Descriptions: We performed a series of comparison and ablation experiments on three real-world datasets obtained from open sources: (1) Venue code scanning data in Chengdu, located in Sichuan Province, China, for a month; (2) Check-in dataset in Tokyo, Japan; and (3) Check-in dataset in New York, USA. Dataset-1 consists of aggregated position data, the entry is formatted as (User ID, Timestamp, Venue Geohash code). Because Database 1 lacks category or semantic information about the visited place, we performed additional work in preprocessing part to satisfy the needs of constructing trajectory graph. Datasets-2 and Datasets-3 (Yang et al., 2015) contain the information on (User ID, POI category name, POI location and Timestamp). Table 1 shows the statistical information of datasets, where Sequences and Graphs refer to the raw trajectories and the constructed spatial-temporal trajectory graphs (ST-Graph), respectively. Figure 3 presents the number of nodes and the maximum outdegree of each graph, both of which follows the power-law distribution.

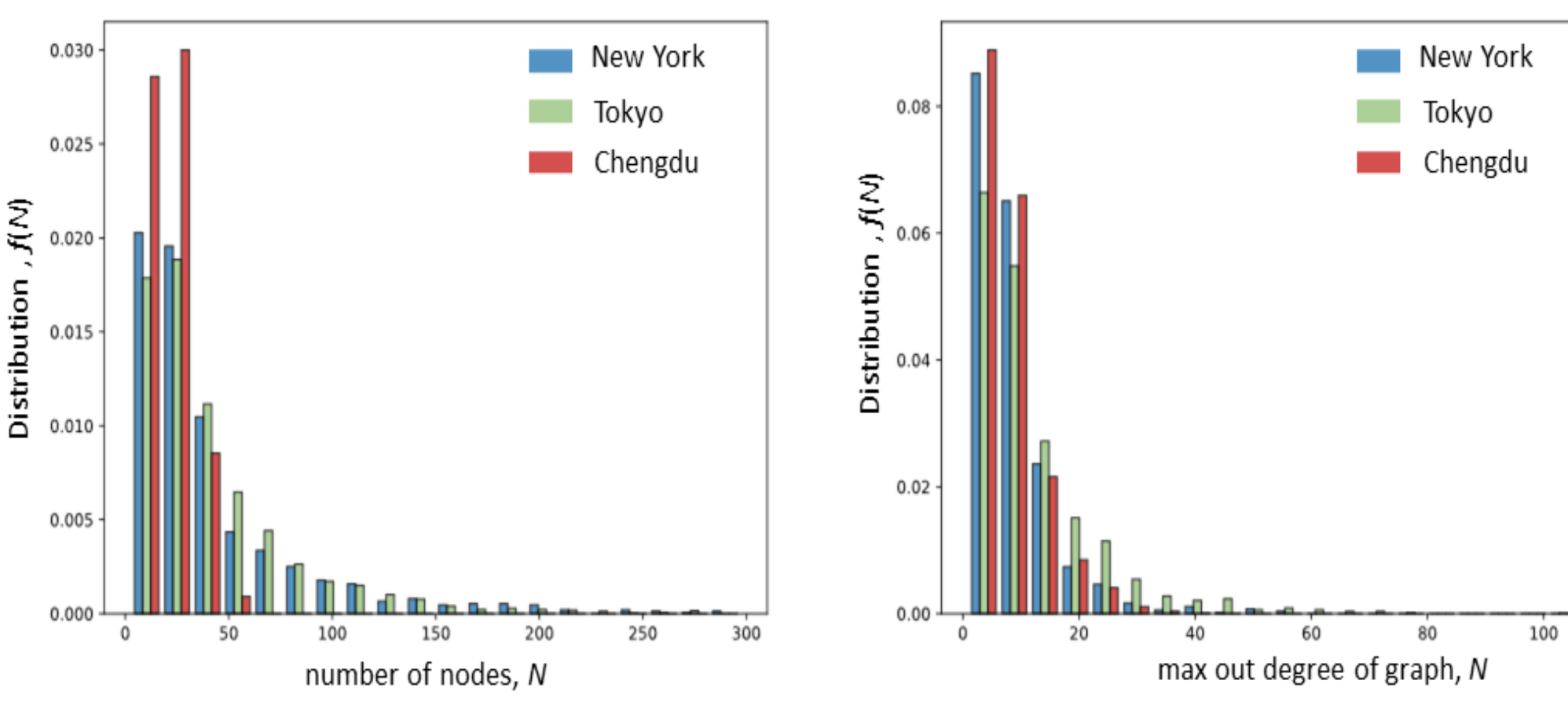

**Figure 3.** Statistics of the trajectory graphs.

**Table 1. Statistics of the experimental data.**

| City | Sequences | Graphs | Edges | POI category | Time Period |
|---|---|---|---|---|---|
| Chengdu | 1,453,219 | 113,109 | 2,319,243 | - | 2021.11.01-2021.11.30 |
| New York | 227,428 | 1,041 | 46,503 | 251 | 2012.04.12-2013.02.16 |
| Tokyo | 573,703 | 2,223 | 93,119 | 247 | |

Data Preprocessing: Before constructing the ST-Graph, the input of ST-GraphRL, we need to preprocess these trajectories. First, to ensure the regularity of the individual mobility, we excluded the individuals with less than three trajectories while ensuring that there are at least two active points in the retained trajectories, i.e., at least one complete mobility transition. The distance, time spent, and frequency of movement were calculated as the weights of movements. Considering the sparsity of the check-in data and the computational overhead, we divide the time interval in the trajectory into 30 minutes, which can be represented through a 48-dimensional vector with start-end time to describe the transitive information of the movement over time. Datasets-2 and Datasets-3 contain 251 and 247 POI names respectively, we encoded them into 128-dimensional feature vectors through Word2Vec to provide more effective semantic information for model. And we summarized them into 10 classes including residential areas, education, food places, transportation, medical, offices/workplaces, personal services, government offices, outdoor and recreation places, and others.

For the Dataset 1, we decode the Geohash to locate the central coordinates (longitude, latitude) of the grid where the event occurred. Then, an preprocessing step involving Graph Neural Networks was devised to learn geo-contextual embeddings by people flow (Z. Liu et al., 2020) to give grids semantic attributes similar to POI categories. Firstly, we leveraged Amap API to collect POI details within each grid, converted POI names into embeddings via Word2Vec, and aggregated them to represent the grid's POI profile. Building upon this, an Origin-Destination (OD) flow network was constructed based on human flow within grids, characterizing nodes with both POI embeddings and daily flow-in and flow-out statistics. The GraphSAGE network (Hamilton et al., 2017) was then employed in a self-supervised learning method to derive 128-dimensional feature vectors for these grids, integrating POI distributions and people flow. At last, we employed Gaussian Mixture model to cluster these features, and the optimal number of clusters were 8 classes. While this approximation may not offer absolutely precise semantic or functional characteristics for each grid, it is a practical solution given the constraints of the available data.

Consequently, nodes in the trajectory graph are uniformly presented as (longitude, latitude, category attributes) where all latitude and longitude values have been standardized. By concatenating the 2-dimenional spatial location with the 128-dimensional POIs or Grids vector, we obtain the 130-dimensional node embedding that integrates geographical and semantic information.

#### 4.1.3 Evaluation Tasks and Metrics

To validate the effectiveness of our ST-GraphRL, three evaluations were conducted: the human mobility distribution prediction, the next location visit prediction, and the Origin-Destination (OD) flow estimation.

**Mobility distribution prediction:** In this task (section 4.2), four evaluation metrics including Accuracy, Precision, and F1 were utilized to verify the prediction for users' spatial-temporal distribution preferences. Specifically, given the recorded data of user $u_i$ to learn the representation $z_i$, and then utilizing $z_i$ to jointly infer time periods in a certain day during which a movement is likely to occur and the category of location the user visited in that time. Meanwhile, suppose there exists a correlation between the mobility pattern and the distribution of the visit frequency, including time and place, coefficient $r$ (Damiani et al., 2020) was employed to evaluate the goodness of the spatial-temporal correlation in embeddings (Section 4.5).

In addition, it is important to evaluate the performance of ST-GraphRL in individual-level trajectory next-location/place prediction(Chu et al., 2024; Rao et al., 2023), in which (section 4.3), MAE, RMSE, MAPE and $\Delta d$ were employed to evaluate the performance of ST-GraphRL in predicting next individual mobility during the weekday and weekend.

**Next Location Visit Prediction:** Given the first location point and time of user mobility on a particular day, combined with the $z_i$ achieved in mobility distribution prediction task as the input, the output is the next possible visited location at the subsequent time point across 24 hours. Based on the predicted results, we calculated the people flow in the grids across 24 hours and compared it with the true number of. Notedly, in the training, the predicted location was presented as the location coordinate ($lat$, $lon$), thus the position error $\Delta d$ refers to the Euclidean distance between the predicted visited location and the truth. In the inferring of people flow, we match the predicted coordinated into the nearest grid such that we can calculate the number of predicted visited people in the place to compare it with true statistic numbers.

**OD Flow Estimation:** We extend the prediction task to estimate both the time and location of user arrivals at the destination, rather than at predetermined hourly intervals. That poses a higher demand for the model to correctly capture spatial-temporal relationships in the user mobilities. The input is the $z_i$ and the given origin's movement time and location. Similarly with the next visit prediction task, the predicted coordinates were match into the grids, then we calculated visited frequency of the OD pairs and validated it with the true OD flows.

Note that, the next visit prediction and the OD flow estimation tasks were conducted on the low-dimensional the representations learned by ST-GraphRL in Dataset-1 for each user. The simple downstream model i.e., a residual network consisting of only two non-linear layers was employed for the prediction of the two tasks, respectively. It is presented as $R_{f,task}(z_i, x_{i,task}) \rightarrow y_{i,task}$, where $R_{f,task}(\cdot)$ denotes the residual network used in the task, $x_{i,task}$ refers to the specified data of user-$i$ in the task, and $y_{i,task}$ is the output.

## 4.2 Individual mobility distribution prediction

### 4.2.1 baseline models

To evaluate the performance of our proposed model in learning spatial-temporal distribution characteristics of the individual mobility, we designed ablation experiments and compare these models with other baseline models.

(1) **Summary trajectory**. The summary trajectory (Damiani et al., 2019) characterizes relevant locations and related mobility patterns in symbolic trajectories, here, symbolic locations are vectorized by word2Vec, and in the experiment N = 3, δ = 2Δ.

(2) **Deep Graph Infomax, DGI**. DGI (Veličković et al., 2018) learns node representations in an unsupervised manner by maximizing mutual information between patch representations and corresponding high-level summaries of graphs. We set the number of layers as 3 in the GNN model.

(4) **Self-attention-based network.** Self-attention (Vaswani et al., 2017) captures context by considering the importance of all elements in the sequence simultaneously, which can model longer-range dependencies effectively than LSTM. We set the hidden layer number as 4 with 8 multi-heads and the hidden channel is 64.

(5) Ablation model **ST-GraphRL-CST**. ST-GraphRL-CST removes the decoupling process from ST-GraphRL, which means that there is no spatial embedding block and temporal embedding block in ST-GraphRL-CST.

(6) Ablation model **ST-GraphRL-NTT**. ST-GraphRL-NTT moves the temporal information into the node feature, meaning that it cuts off the edge attribute.

(7) Ablation model **ST-GraphRL-DGI**. ST-GraphRL-DGI replaces the decoupling process used in ST-GraphRL with the linking prediction employed in DGI, such that comparing which kind of initial embedding over space and time is better to learn the coupled spatial-temporal dependency.

In summary, the Summary trajectory is a non-deep learning method specifically designed for trajectory representation considering location information; the Self-attention-based model performs well in extracting series relationships, and DGI enables capturing spatial relationships between movements. To ensure the equality of experiments, the decoder designed in ST-GraphRL was also added to baseline models for training. The Chengdu dataset was divided into two sets: 80% of the graphs were the training set and the remaining 20% graphs were the testing set. Since the sizes of the New York dataset and Tokyo dataset are small, we did not split the training set. All the evaluations were performed on a ×64 machine with a single GPU (NVIDIA RTX 1080Ti).

**Table 2. Evaluation of the mobility distribution prediction task.**

| Method | Datasets | Accuracy ↑ | Precision ↑ | F1 ↑ |
|---|---|---|---|---|
| Summary Trajectory | New York | 0.2983 | 0.1978 | 0.2320 |
| | Tokyo | 0.3075 | 0.2818 | 0.1670 |
| | Chengdu | 0.2656 | 0.3465 | 0.2016 |
| DGI | New York | 0.4337 | 0.4117 | 0.3954 |
| | Tokyo | 0.4947 | 0.4569 | 0.4616 |
| | Chengdu | 0.5076 | 0.4765 | 0.4402 |
| Self-attention | New York | 0.5406 | 0.5274 | 0.5192 |
| | Tokyo | 0.5772 | 0.5272 | 0.5104 |
| | Chengdu | 0.6716 | 0.5603 | 0.5684 |
| ST-GraphRL-NTT | New York | 0.5735 | 0.5562 | 0.5460 |
| | Tokyo | 0.5959 | 0.5530 | 0.5640 |
| | Chengdu | 0.6527 | 0.5541 | 0.5537 |
| ST-GraphRL-CST | New York | 0.5335 | 0.4915 | 0.5130 |
| | Tokyo | 0.5780 | 0.5135 | 0.5210 |
| | Chengdu | 0.6787 | 0.5615 | 0.5334 |
| ST-GraphRL-DGI | New York | 0.5602 | 0.5418 | 0.5225 |
| | Tokyo | 0.6081 | 0.6035 | 0.5560 |
| | Chengdu | 0.6750 | 0.5653 | 0.5470 |
| ST-GraphRL | New York | 0.7064 | 0.6302 | 0.6052 |
| | Tokyo | 0.7739 | 0.6771 | 0.6548 |
| | Chengdu | 0.8540 | 0.6451 | 0.6311 |

### 4.2.1 Comparison of prediction results

We compared our method with the baseline methods in the task of imitating the spatial-temporal distribution of individual trajectories in one day. The evaluation metrics including Accuracy, Precision, and F1 are listed in Table 2. It is found that the Summary trajectory presented the lowest scores in all metrics, which may be because it cannot interpret trajectory similarity in terms of location similarity; moreover, its sequential structure is hard to deal with the temporal dimension of sparse trajectories. DGI outperformed the Summary trajectory with increased average score ($\Delta accuracy: +0.19, \Delta precision: +0.17, \Delta F1: +0.23$) in three datasets, which demonstrates that the presentation of node-linking-node in the graph is effective to represent movements. However, F1 scores achieved by DGI on all datasets are still less than 0.5, indicating that it failed to capture movement patterns. The Self-attention achieved higher scores than DGI ($\Delta accuracy$: $+0.12$, $\Delta precision$: $+0.09$, $\Delta F1$: $+0.10$), but its average F1 score is only 0.03 higher than 0.5. It indicates that although the Self-attention can learn contexts from sparse features, while it can only capture partial, in certain time periods, mobility patterns. Our proposed method, ST-GraphRL, as expected, achieved the best performance with more than 0.6 scores in accuracy, precision, and F1 over both three datasets, demonstrating the effectiveness of joint spatial-temporal representation.

Detailly, F1 scores of three ablation models, ST-GraphRL-CST, ST-GraphRL-NTT, and ST-GraphRL-DGI, fluctuated between 0.50-0.56, demonstrating the necessity of the components in ST-GraphRL. Compared to ST-GraphRL-NTT which discards the presentation specializing in temporal information, removing the step of decoupling result in a 0.2 decrease in F1 scores to ST-GraphRL-CST, suggesting that decoupling plays a vital role in clearing up entangled spatial-temporal relationships. The Self-attention performed comparably to ST-GraphRL-NTT on the Chengdu dataset, but scored lower than ST-GraphRL-NTT on the other two datasets. It implies that larger datasets may improve Self-attention performance but that its sequential structure still confounds spatial-temporal dependence when temporal transitions between locations are not explicitly presented. When we retain the edge attribute used in ST-GraphRL, while replacing the decoupling strategy with linking prediction used in DGI, the ST-GraphRL-DGI achieved similar scores to that of ST-GraphRL-NTT, with improved performance compared with DGI. In summary, the results prove that joint encoding is the basis for accurately learning complex spatial-temporal dependencies in the moving process. It presupposes that the transfer relation in trajectories over space and time is explicitly constructed and that the decoupling-fusion strategy is an effective method for achieving efficient spatial-temporal joint modeling.

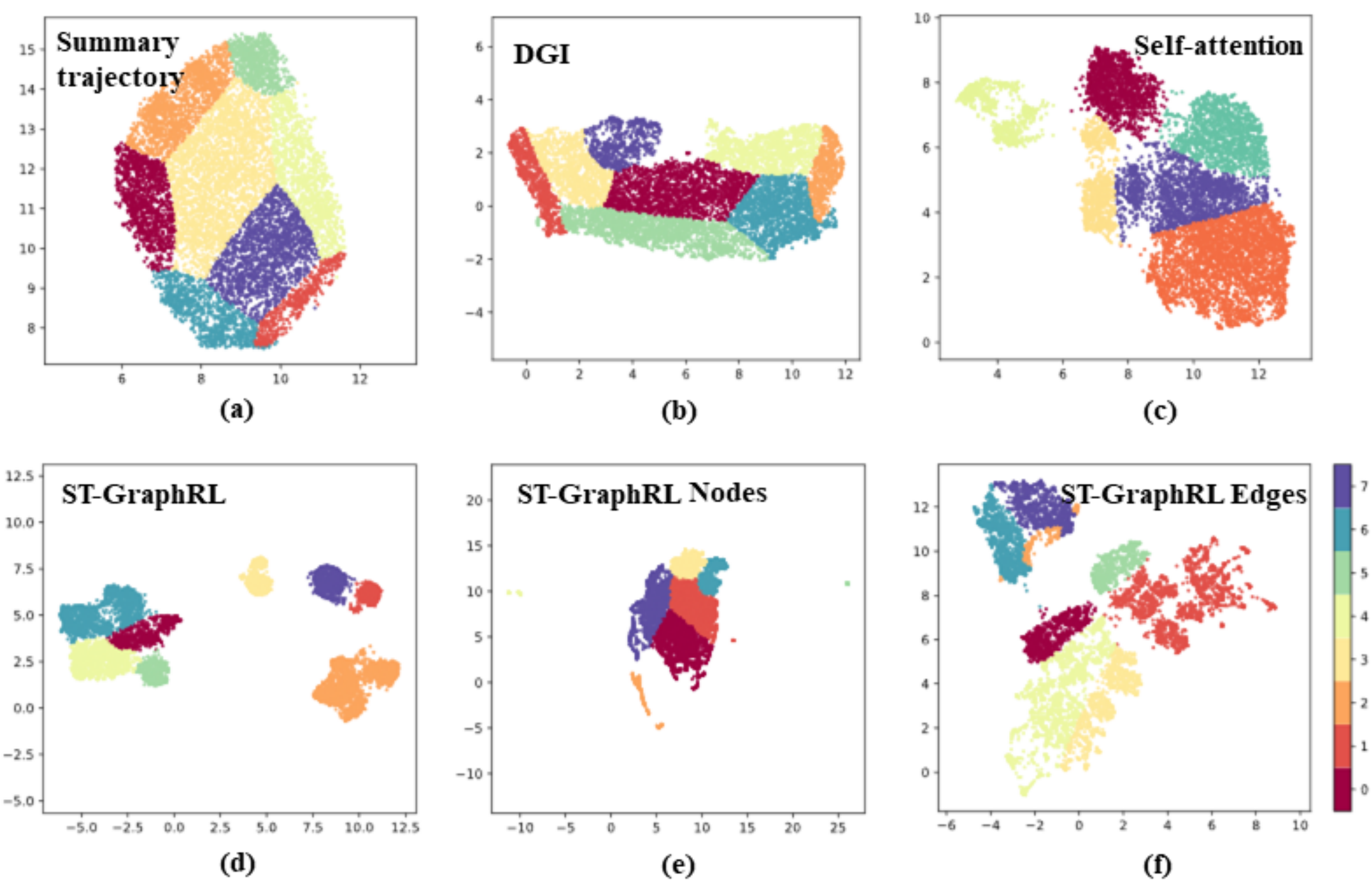

**Figure 4.** Dimensional reduction of trajectory representations with 8 clusters; Nodes refer to the vertex features of ST-GraphRL; Edges refer to the combination of one edge feature and two vertex features connected by the edge.

### 4.2.2 Visualization of embeddings

To further validate that ST-GraphRL can distinguish trajectory similarities, we reduced and visualized the representation space to 2-dimensions by UMAP. In Figure 4, the colored points are the clustered individuals. The spaces among each cluster in ST-GraphRL (Figure 4(d)), are much larger, such that clustersan easily be separated visually even without the color reference, which it failed to achieve in the Summary trajectory (Figure 4(a)), DGI (Figure 4(b)), and Self-attention (Figure 4(c)). Furthermore, we visualize the embedding of nodes and edges (node pairs) in ST-GraphRL (Figure 4(e-f)). It is found that edges present a comparable performance with the Self-attention, their embeddings are not particularly tightly clustered in single clusters, but there is comparably distinguishability between different clusters. It indicates that the features and patterns they captured may also be similar. Although there is compacted node embedding, ST-GraphRL shows better separability of its inter-cluster than that of edges and the Self-attention. It confirms that features learned in nodes and edges are different, focusing on independent patterns in the latent feature space, while fusing the two kinds of features helps to completely represent a movement. In contrast, the Self-attention can only capture partial features, i.e., the similarity between different trajectories rather than mobility patterns, as it cannot learn the interactive relationships over space and time hidden in trajectories. In summary, representations generated by the proposed ST-GraphRL can exactly measure the differences and similarities in individual trajectories, which is attributed to the accurate representation of both spatial and temporal dimensions by its nodes and edges, as well as the modeling of the relationships between nodes and edges.

### 4.3. Next location visit and OD flow prediction

Based on the results achieved in Section 4.2, this section compared the ST-GraphRL with the Self-attention based model and the MLP-LSTM based model from (Rao et al., 2020) to evaluate the performance of ST-GraphRL in downstream tasks including the next visited location prediction and OD flow estimation. The selected MLP-LSTM is a suitable model for comparison due to its decoupled spatiotemporal information design concept.

**Table.3 Prediction results and comparison with baselines.**

| Method | Period | OD Pairs | | | People Flow | | | Next location |
|---|---|---|---|---|---|---|---|---|
| | | MAE↓ | RMSE↓ | MAPE↓ | MAE↓ | RMSE↓ | MAPE↓ | $\Delta d$(km)↓ |
| Self-attention | | 4.7348 | 17.1111 | 0.8799 | 23.8051 | 24.9988 | 0.0219 | 0.6558 |
| MLP-LSTM | weekdays | 4.2252 | 28.1354 | **0.4903** | 21.9155 | 22.9690 | 0.0201 | 0.4435 |
| ST-GraphRL | | **4.0031** | **8.5216** | 0.5118 | **9.9769** | **11.7082** | **0.0107** | **0.1543** |
| Self-attention | | 2.0863 | 5.1092 | 0.5679 | 11.5453 | 15.0235 | **0.0131** | 0.3191 |
| MLP-LSTM | weekends | 1.6002 | 5.1928 | 0.4069 | 15.1513 | 16.7526 | 0.1033 | 0.3444 |
| ST-GraphRL | | **0.7006** | **3.3831** | **0.1647** | **2.9532** | **4.2541** | 0.0126 | **0.1316** |

#### 4.3.1 Comparison of prediction results

Table.3 reports the ST-GrpahRL's performance in OD flow estimation and people flow estimation based on the next visit prediction during weekdays and weekends. It shows that ST-GraphRL achieved the lowest MAE 4.0031 and RMSE 8.5216 in the OD pairs prediction during the weekdays, in which the MLP-LSTM had much higher RMSE 28.1354. The MLP-LSTM and the self-attention performed well in MAPE, 0.4903 and 0.0131, respectively in the pair OD (on weekdays) and people flow (on weekends) estimation task, while ST-GraphRL exhibited a remarkably close performance to them in this aspect. ST-GrpahRL obtained the best performance in the next visited location prediction, decreasing errors of the MLP-LSTM from 0.4435 to 0.1543, and 0.3444 to 0.1316, respectively. Moreover, it is found that ST-GraphRL presented consistently excellent performance on the tasks both on weekends and weekdays. That indicates ST-GraphRL enables to infer weekly mobility patterns based on the achieved hourly user representation that contains accurately captured spatial-temporal dependency.

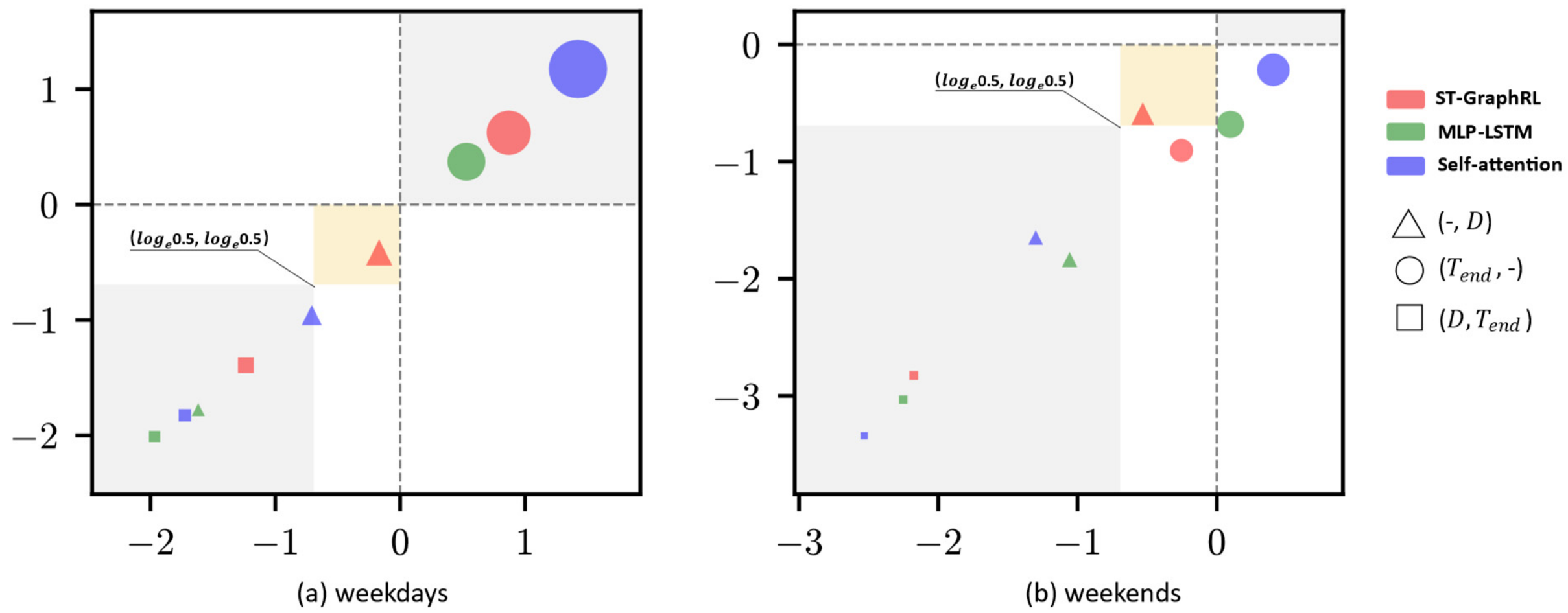

**Figure5.** Evaluations of model's performance on the OD flow prediction from the space and time scale. the yellow zone indicates the high performance, the grey zone in the upper right and the lower left refers to an overfitting performance and underfitting performance of models in the task, respectively.

Furtherly, we analyze OD pairs predicted correctly in three aspects: Given the original grid at the departure time, matching with the arrival time, the destination grid, and the destination grid at arrival time i.e., (O, $T_{start}$)→[($T_{end}$, -), (-, $D$), ($D, T_{end}$)]. Then, the matched (O, $T_{start}$)→($T_{end}$, -), (O, $T_{start}$)→(-, $D$), and (O, $T_{start}$)→($D, T_{end}$) were counted up and calculated to obtain the percentage of them in the whole predicted OD flow; such that we can comprehensively evaluate the model's propensity and performance on spatial-temporal prediction downstream tasks. Figure.6 reports the further evaluations on the predicted OD flow, in which the scatters = ($log(p^{i,j}_{true})$, $log(p^{i,j}_{pre})$), where $i \in$ {ST-GraphRL, MLP-LSTM, Self-attention}, and $j \in$ {($T_{end}$, -), (-, $D$), ($D, T_{end}$)}; and $p^j = (n_j/N)$ where $n_j$ refers to the number of matched OD flows, and $N$ is the total OD flows. Notedly, scatters of (O, $T_{start}$)→($T_{end}$, -) are presented gray zone at the upper right corner of Figure5. It means all the models tend to predict frequent start-end time pairs, but often with significant deviations in predicted arrival times compared to actual arrival times, leading to $log(p^j)$>1, especially to the self-attention model. In Figures5, the red scatter points of ST-GraphRL distributed inside or around the yellow zone, indicating that ST-GraphRL predicts OD flow that closely resembles the true distribution of the OD flow. Particularly in the destination prediction (O, $T_{start}$)→(-, $D$), ST-GraphRL is the only one model in the yellow zone, and nearly perfectly predicts the ground truth. While there is still room for improvement in the state considering both time and space $(O, T_{start}) \rightarrow (D, T_{end})$, the performance of ST-GraphRL is significantly better than that of self-attention and MLP-LSTM. Figure 6 displays the OD flow in selected regions, where the variation of the color refers to arrival time, and line thickness reflects the frequency of OD flow (only visualizing OD flows with frequency greater than 10). These results confirm that ST-GraphRL can effectively predict the spatiotemporal distribution of OD flow, whether in dense or sparse OD flow areas.

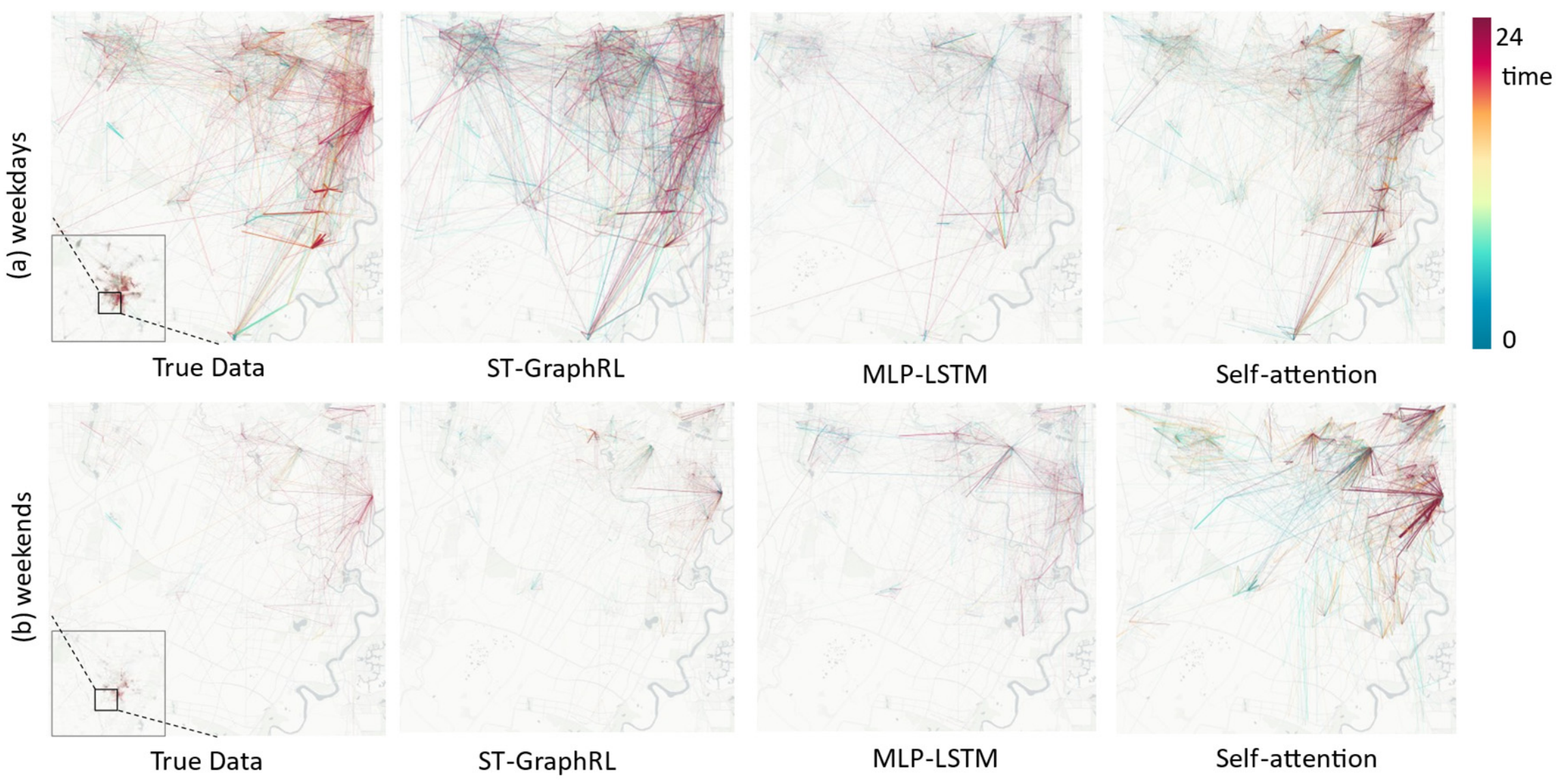

**Figure 6.** A visual comparison of the OD flow predicted by models.

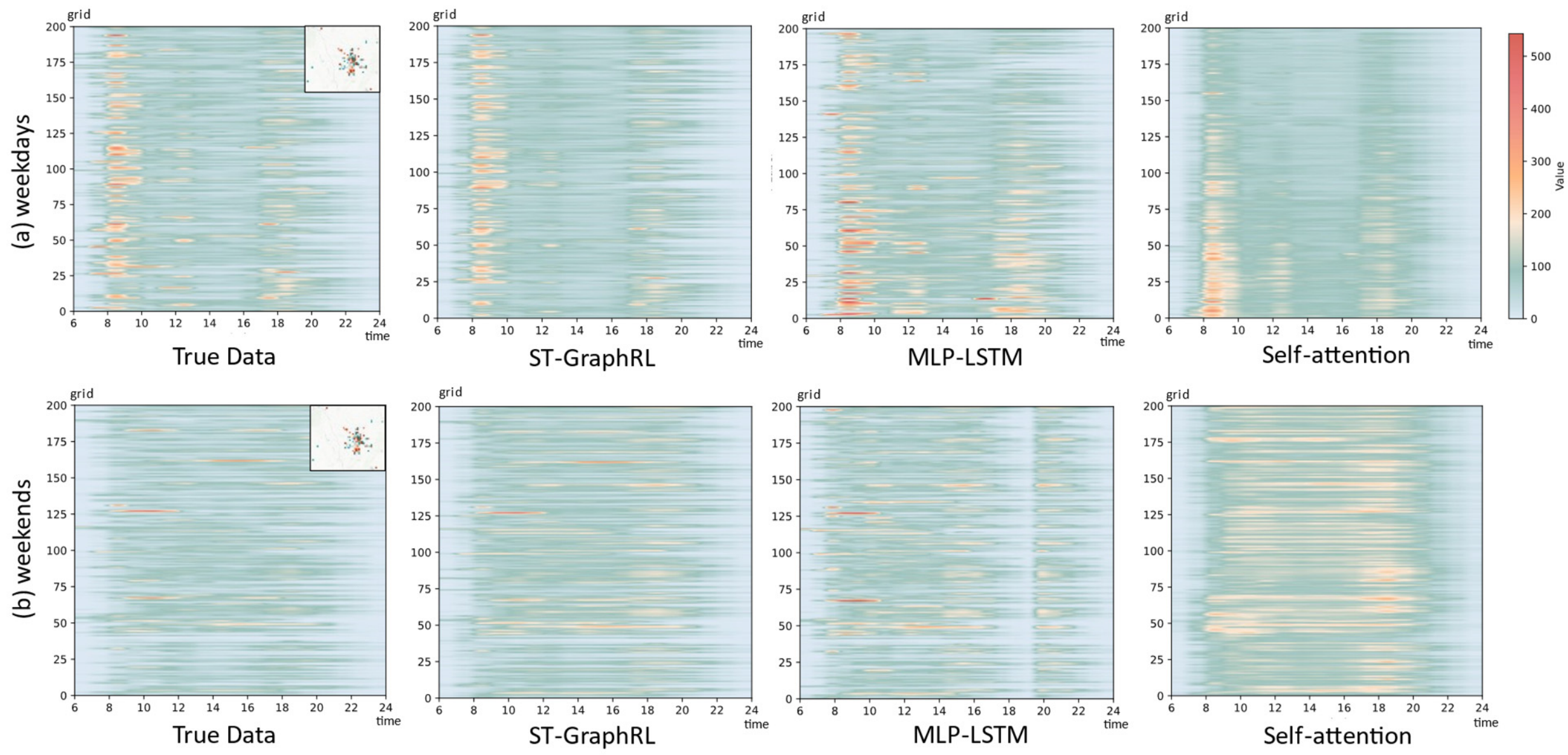

**Figure 7.** The temporal distribution of the people flow in grids.

Figure.7 reports the people flow in randomly selected 200 grids from 6:00 AM to midnight, based on the next location visit prediction task. It is observed that the results of ST-GraphRL align most closely with the actual situation, while MLP-LSTM and self-attention models tended to overestimate the number of people. Combined with Figure 6, the self-attention highly exaggerated the people flow in some grids, indicating that it might just remember the shallow features such as high traffic areas instead of learning the spatial interaction and pattern in individuals' mobility. Deriving benefit from the spatial-temporal graph and the jointly encoding over space and time, ST-GraphRL enables to accurately capture spatial-temporal dependency that is important to the OD flow prediction. Combined with Figure 8, the self-attention model had the highest errors on both weekdays and weekends, while MLP-LSTM performed better on weekends than weekdays but still exhibits sensitivity to specific time periods i.e., particularly with higher prediction errors around 8:00 AM and 11:00 AM. That might be because although MLP-LSTM designed with the decoupled spatiotemporal representation learning structure, integrating and model the interaction across the space and time scale in mobility is difficult to the simple MLP blocks, especially, to deal with sequential trajectories. By contrast, ST-GraphRL achieved the greatest overall performance, demonstrating both the superiority of the ST-GraphRL structure and ST-Graph that supports our model to learn the spatial-temporal patterns of individual movement.

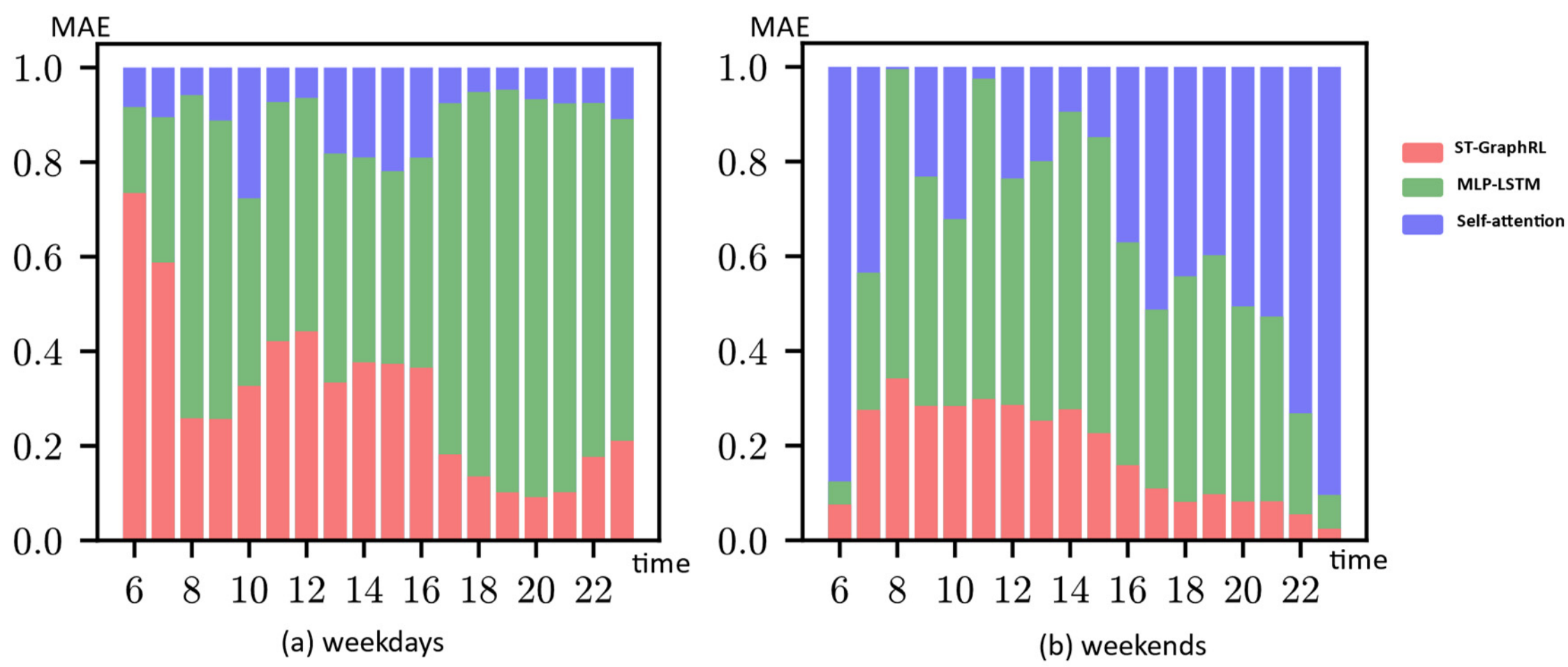

**Figure 8.** Comparison of the estimated people flow errors in grids of three models. the errors of each model in different time intervals were normalized by dividing the maximum MAE among the three models in that particular time interval. This normalization process ensured that the error values were scaled to a range between 0 and 1.

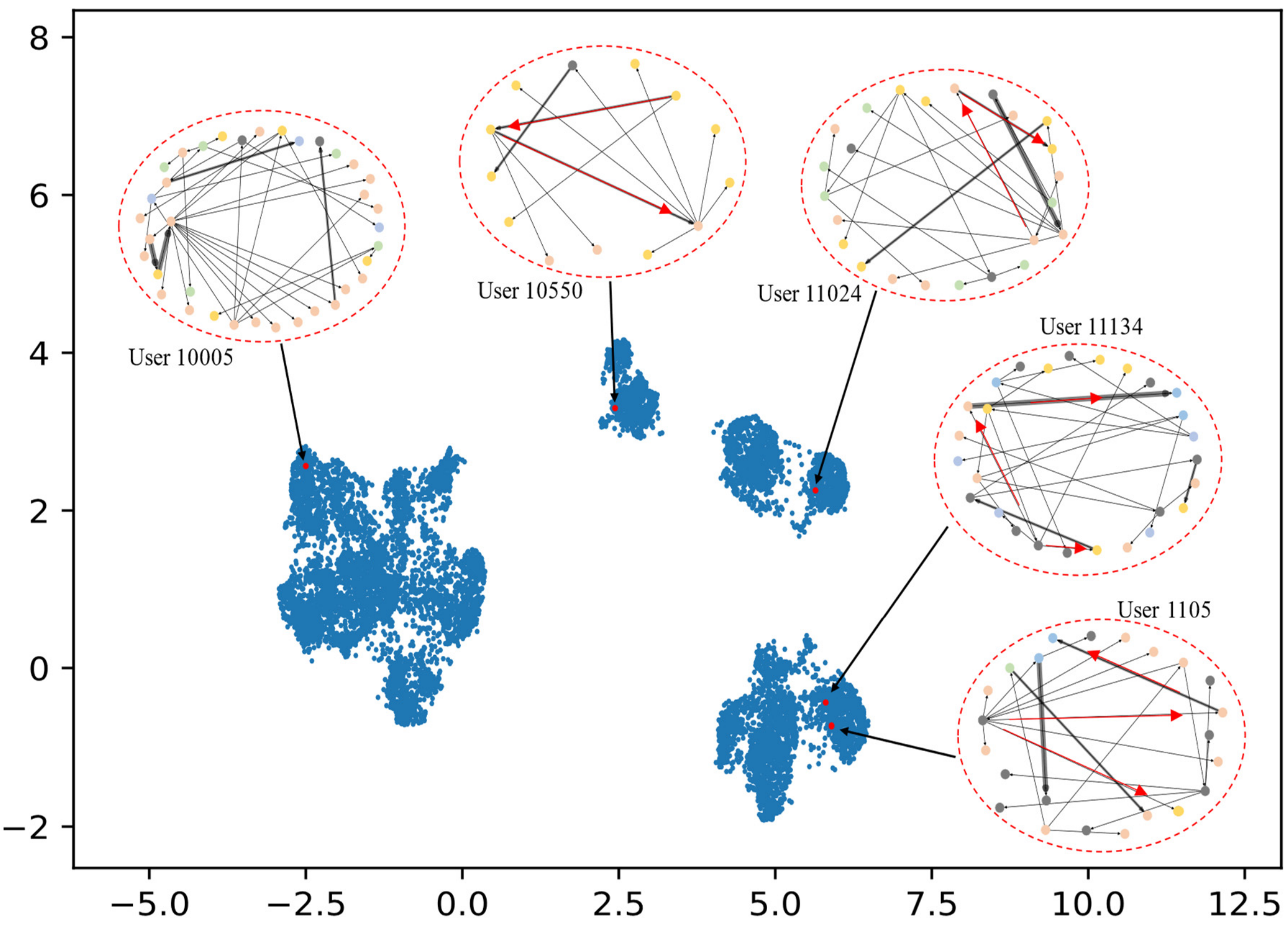

**Figure 9.** Trajectory graphs corresponding to embeddings represented by ST-GraphRL.

### 4.4 Visualization of trajectory graph

We visualized the trajectory graph for some randomized individuals to evaluate the distance among the respective representations. As seen in Figure 9, the color of nodes indicates POI's category, the arrow refers to the order of visits, and the thickness of the line represents the frequency of visits; in which graphs of individuals in different clusters show differences in the following aspects: the number of nodes, the category of nodes and node pairs, and the frequency of visits. As marked with red arrows in user 11134 and user 1105, they show similar movement sequences and visited frequency between two nodes, thus they are close in the clustering space. Also, the diversity, in terms of space and time, of trajectories can be reflected on the trajectory graph. Such as, user 10550 shows great regularity with fixed types and high frequencies of places to visit, while the movements of user 10005 are more random in temporal dimension as there are more node pairs with the same color. Although user 11024 and user 10550 have similar visited locations, the order of their visits is very different, e.g., user 11024 takes the path "yellow point → yellow point → pink point", while user 10550 takes the path " pink point → pink point → yellow point", thus they are not in the same cluster.

### 4.5 Validation of spatial-temporal dependency in representations

To further validate whether ST-GraphRL enables the capture of spatial-temporal dependency in trajectories, we employed the following approach: assuming that there is a specific spatial-temporal mobility pattern in each individual's trajectories, and thus the similarity of mobility patterns between different individuals can be measured by calculating the distance ($d_{true}$) of the movement spatio-temporal distribution matrix. If the distance between the trajectory representations ($d_{rep}$) of individuals has a positive correlation ($r$) with $d_{true}$, it suggests that the learned representation already encompasses the spatial-temporal dependency in trajectories and is representative of the individual mobility pattern. The Euclidean distance and the Jensen distance are utilized to calculate for $d_{rep}$ and $d_{true}$. We measured the similarity of individual trajectories from three indicators, i.e., the distribution of categories of visited locations, the distribution of times when the movement occurred, and the spatial-temporal distribution of the movement, denoted by $r_s$, $r_t$, and $r_{st}$, respectively.

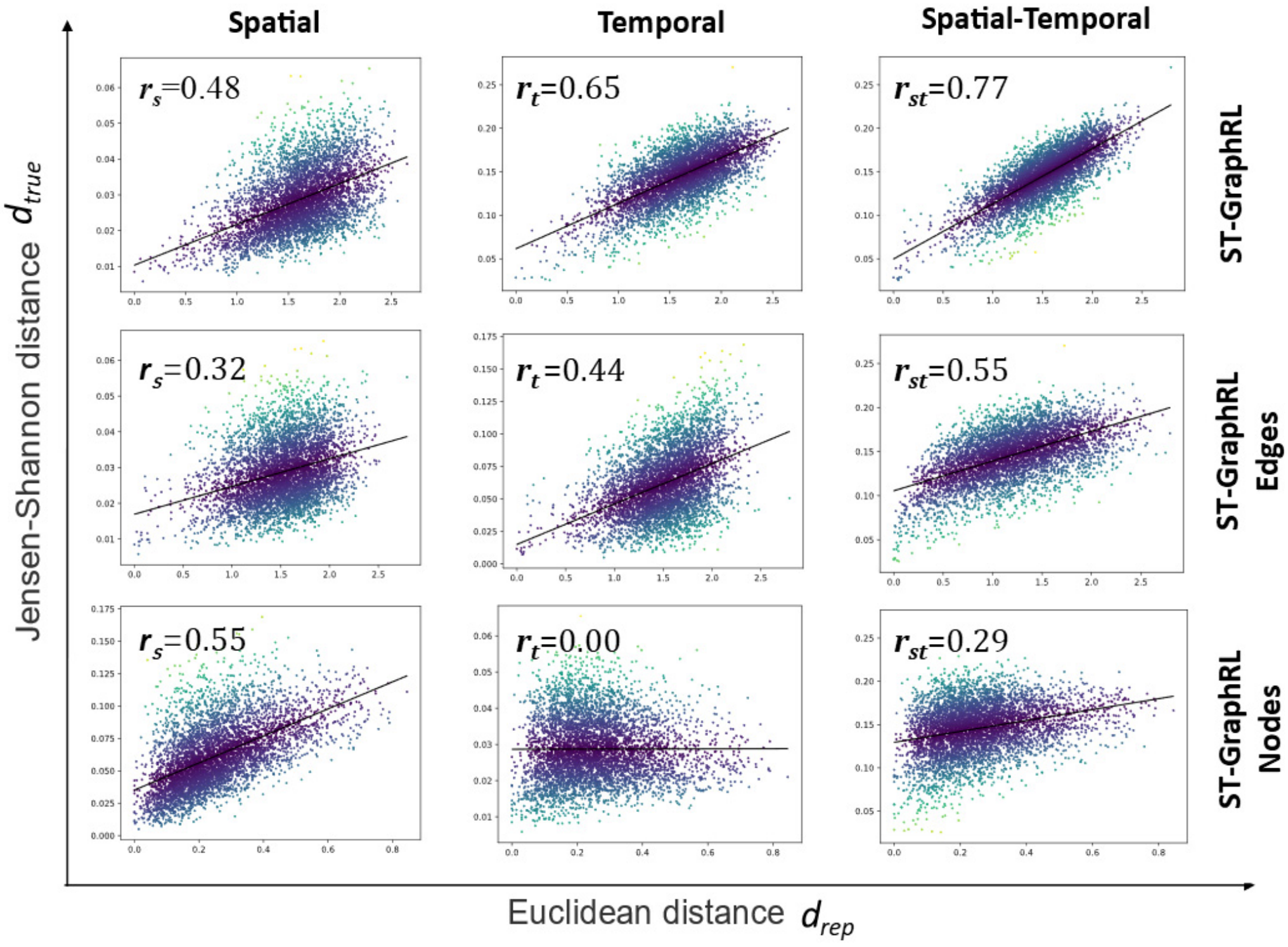

**Figure 10.** The correlations between the true distributions of trajectories and the similarity of individuals' embeddings learned by ST-GraphRL in the Chengdu dataset: the x-axis reports Euclidean distance between embeddings, the y-axis refers to Jensen distance of distributions of true trajectories.

**Table 4. Evaluation of the spatial-temporal dependency in embeddings in Chengdu, Tokyo, and New York dataset (CD, TKY, and NYK); $r_s$, $r_t$, and $r_{st}$, respectively refers to the correlation with the distribution of location frequency, visited time-period frequency, and spatial-temporal frequency in trajectories. the higher the value of $r$, the better the performance of the model.**

| Method | $r_s$ ↑ | $r_t$ ↑ | $r_{st}$ ↑ |
|---|---|---|---|
| | CD/ TKY/ NYK | CD/ TKY/ NYK | CD/ TKY/ NYK |
| ST-GraphRL | 0.48/0.35/0.04 | **0.65**/0.16/0.15 | **0.77**/0.48/0.18 |
| ST-GraphRL (Nodes) | **0.55**/0.55/0.02 | 0.00/0.17/0.06 | 0.29/0.39/0.16 |
| ST-GraphRL (Edges) | 0.32/0.18/0.09 | 0.44/0.28/0.11 | 0.55/0.40/0.38 |
| DGI | 0.24/0.16/0.04 | 0.33/0.12/0.06 | 0.26/0.12/0.02 |
| Summary trajectory | 0.05/0.03/0.01 | 0.03/0.06/0.01 | 0.03/0.00/0.03 |
| Self-attention | 0.31/0.12/0.04 | 0.56/0.16/0.12 | 0.58/0.31/0.12 |

As displayed in Table 4 and Figure 10, the representation of ST-GraphRL exhibits a strong coherence with the spatial-temporal distribution of trajectories, particularly on the Chengdu dataset with a coefficient $r_{st}$ 0.77. It confirms the existence of abstracted spatial-temporal dependency features in the latent space of ST-GraphRL representations. In contrast, the Summary Trajectory and DGI fail to capture spatial-temporal distribution, showing a negligible correlation $r_{st}$ 0.03 and very weak correlation $r_{st}$ 0.26 between their representation and the actual trajectory distribution, respectively, leading to their low scores reported in Table 2. The Self-attention excels at extracting relations between consecutive contexts, providing robust support for it to learn trajectories' temporal distribution $r_t$ 0.56. However, the Self-attention ignores the features in movement transitions, resulting in decreased scores $\Delta r_s = -0.17$ and $\Delta r_{st} = -0.19$ compared to ST-GraphRL in the Chengdu dataset. The embeddings of nodes and edges in ST-GraphRL exhibit correlations with the distribution of trajectories in space and time, respectively. The node achieves the highest $r_s$ 0.55, the edge has the second highest $r_t$ 0.44 and $r_{st}$ 0.55, indicating that the node and edge in ST-GraphRL effectively represent the spatial and temporal features, respectively. Thus, integrating features from both nodes and edges enables ST-GraphRL to ultimately achieve the highest correlation $r_t$ 0.65 and $r_{st}$ 0.77.

The result validates that explicit structuring and decoupling of space and time (i.e., the edge and node) helps the extraction of accurate spatial and temporal features from entangled information; and the jointly features fusing empowers ST-GraphRL in effectively learning spatial-temporal dependencies and distributions.

## 5. Discussions: neural responses to spatial-temporal patterns in latent space

The above results have demonstrated the ability of ST-GraphRL in representing space and time features. But how can the ST-GraphRL express spatial-temporal interactions/patterns in trajectories? To answer the question, we further explore how the activated representation in latent space responds to spatial-temporal patterns that are described by two specific designed spatial-temporal indexes, $I_{st}^1$ and $I_{st}^2$. Noting that, an effective response is defined as the observable and regular variation (e.g., a linear variation) of response strength to respond the changes of indexes:

$$I_{st}^1 = \sqrt{{n_t}^2 + {n_s}^2} \quad (13)$$

where $n_t$ refers to the average daily number of movements, while $n_s$ refers to the average number of different categories of places the user visits, here $n_t$ and $n_s$ are normalized. The higher $I_{st}^1$, the greater spatial-temporal diversity of an individual's mobility.

$$I_{st}^2 = \log\left(\frac{\sum_N^i d_i}{N}\right) \quad (14)$$

where $\boldsymbol{d_i}$ refers to the cosine similarity between the *i*-th movement and other *N*-i movements, N is the total number of moves made by an individual. Here, $I_{st}^2$ is employed to evaluate the stability of mobilities over space and time. The higher the $I_{st}^2$, the greater the possibility of fluctuations in a regular mobility pattern, e.g., assuming the individual moves many times, there may be the same departure-arrival location, but the start-end time is different; or there is the same start-end time, but the departure or arrival locations are different.

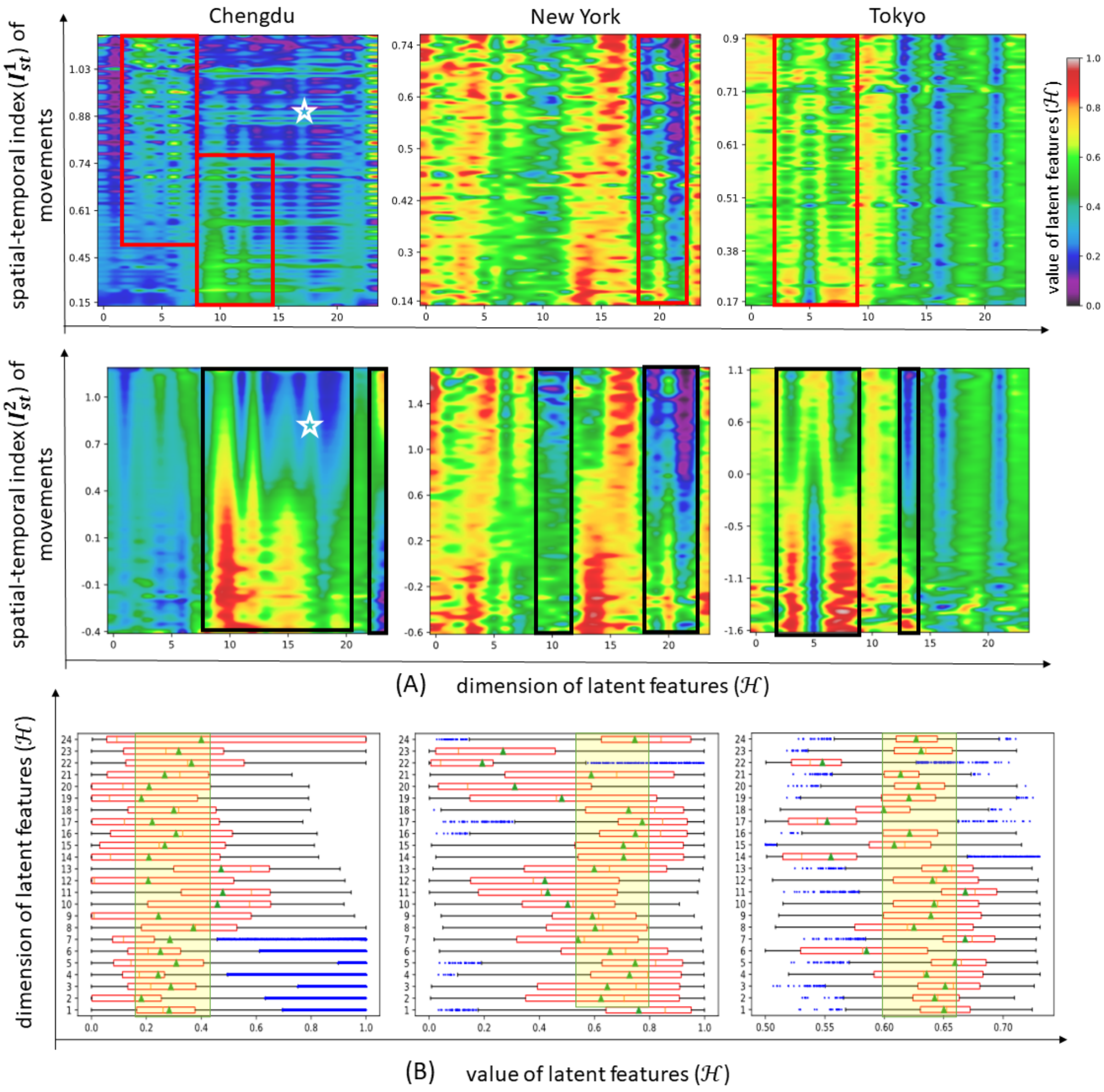

**Figure 11.** Spatial-temporal patterns represented in latent space. (A) refers to heat maps of neuronal response to spatial-temporal patterns; (B) refers to the statistical distribution of feature value of individual trajectory representations in latent space.

As shown in Figure 11-(A)-Chengdu, strong/weak response zones with distinct boundaries are formed in the latent feature space, and the response strength varies linearly in a specific range of dimensions. Detailly, in dimensions 2-7, we observed that the response intensified with the increasing $I_{st}^1$; conersely, in dimensions 7-14, responses weakened with higher values of $I_{st}^1$. Showing that the neuronal responses in the two spans have a positive/negative linear relationship with the change of the spatial-temporal pattern defined by $I_{st}^1$. Notably, the responses to the pattern described by $I_{st}^2$ is stronger and more pronounced. In dimensions 7-20, it is noted a reduction in neuronal responses as $I_{st}^2$ increased, while in dimensions 22-24, the response strengthened with increased $I_{st}^2$. It is found that activated dimensional spans to $I_{st}^1$ and $I_{st}^2$ are partially similar, while the strength of their responses is differed significantly, especially in the regions around with the white star. This indicates that there are specified responses to space and time, and different compositions of these responses can represent different kinds of spatial-temporal patterns. For instance, when the target pattern changes from $I_{st}^1$ to $I_{st}^2$, there is observed a shift in the strong response interval from dimensions 2-14 to dimensions 7-20. This suggests that latent representations of ST-GraphRL adapt and represent different spatial-temporal patterns in a structured manner.

On the New York and Tokyo datasets, where neurons were generally in a high activation, responses to specific spatial-temporal patterns were not much easy to bind, but some regularities in these responses can still be detected. In the New York dataset, there are similar negative linear responses to both $I_{st}^1$ and $I_{st}^2$, while the $I_{st}^2$'s response is clearer and more continuous; moreover, a new response slightly emerges in dimensions 9-11 in $I_{st}^2$. In the Tokyo, the positive and negative response patterns were shown in $I_{st}^1$ and $I_{st}^2$ in dimension-9, respectively. And a significant negative linear response to $I_{st}^2$ appeared in dimension-14. By analysing responses heat maps in all three datasets, it confirms that features learned in latent space enable to capture different spatial-temporal interactions by producing and compositing the corresponding responses.

We visualized the statistical information of representations to further study the reasons for the diverse response heat maps presented on the three datasets, which have uneven data size and individuals in them may have different moving patterns. The boxplot in Figure 11-(B) presents the distribution of the feature values of representations in 24 dimensions. It is found that the distributions in New York and Tokyo are similar, their outliers are dispersed, indicating that the neurons would be more active while less responsible for a certain response so that the representation is less sensitive to capture spatial-temporal patterns. In contrast, the feature values of the Chengdu dataset are polarized while tightly distributed, e.g., the outliers are concentrated in dimensions 1-7. Additionally, the span of distribution mean, marked with green triangles, in New York and Tokyo is concentrated at 0.5-0.8 and 0.6-0.66, respectively, while that of Chengdu is around 0.1-0.3; callback to Figure 11-(A), the dimensions whose mean value deviating from this span would be activated and responds to designated spatial-temporal patterns. It indicates that the larger training dataset helps ST-GraphRL to converge in a narrower latent space with a more compact feature distribution (Chung et al., 2022), as well as showing weaker neuronal sensitivity, i.e., only a small number of neurons would be activated when exposed to a specific spatial-temporal pattern, which is consistent with the response pattern of neurons in the human brain.

In summary, the linear response presented in latent space implies that the spatial-temporal patterns can be explicitly observed and represented in the representation vectors generated by ST-GraphRL; in other words, ST-GraphRL successfully learns spatial-temporal interactions in individuals' mobility and extracts patterns from these interaction features.

## 6. Conclusions

Spatial-temporal dependency is crucial for comprehending geospatial phenomena, making it essential for advancing the development of human-centric GeoFMs. However, the sparsity and high-dimensional coupling of spatial-temporal features within individual trajectories pose challenges in obtaining general-purpose trajectory representations capturing the spatial-temporal dependencies. To address this, this paper proposes a spatial-temporal joint representation learning method for individual trajectories. Experimental results on three real-world mobility datasets and investigation to the latent feature space validate the effectiveness of ST-GraphRL. Representations generated by ST-GraphRL were evaluated using multiple metrics, whose advantage is seen in its accurate prediction to the spatial-temporal distribution movements and measurement of trajectory similarity, the tasks that other comparison approaches find challenging.

ST-GraphRL's superior performance is attributed to two novel designs: (i) the spatial-temporal trajectory graph employs the temporal transition vector as a directed bridge linking spatial features. The explicit and structured presentation of space and time information can effectively record spatial-temporal interactions in movements; (ii) the two-stage jointly encoding model, including decoupling and fusing: independently encoding to establish initial embeddings of time and space aids in disentangling intricate spatial-temporal interactions; then message passing and aggregating operations joint space and time features facilitates the learning of spatial-temporal dependencies. In final, the decoder, promoting the representations to statistically simulate spatial-temporal distributions of mobilities, which contributes to understand mobility patterns.

The proposed ST-GraphRL may also have implications for representation learnings of other geospatial data. In this particular instance, the spatial-temporal dependence of human mobility could be considered. This implication is particularly crucial for the advancement of GeoFMs, as their effectiveness may rely significantly on their ability to acquire Geo-knowledge and identify patterns concealed within various geospatial data sources. Our work not only shed light on the structurally-explicit spatial-temporal relationships in representation learning, but it also has implications for social and ethical considerations. For instance, the well-represented relationships cross space and time scale will support deep learning models to synthesize high-quality human trajectory data in various urban systems, which can be used to address "data inequity" and "data privacy" issues in related studies. Furthermore, comprehending the explicit spatial-temporal relationships can enhance our understanding of the intricate dynamics between humans and urban environments.

## Author statement

## Acknowledgments

## Notes on contributors

## Data and codes availability statement

The Check-in dataset in Tokyo and New York that support the findings of this study were derived from the following resources available in the public domain: [https://privamov.github.io/accio/docs/datasets.html]; the dataset in Chengdu cannot be shared publicly due to restrictions.